\documentclass[journal]{IEEEtran}
\usepackage{graphicx}
\usepackage{booktabs}
\usepackage{times}
\usepackage{soul}
\usepackage{color}
\usepackage{xcolor}
\usepackage{url}
\usepackage{amsfonts,amssymb}
\usepackage[hidelinks]{hyperref}
\usepackage[utf8]{inputenc}
\usepackage[small]{caption}
\usepackage{graphicx}
\usepackage{amsmath}
\usepackage{amsthm}
\usepackage{booktabs}
\usepackage{algorithm}
\usepackage{xcolor} 
\usepackage{algorithmic}
\usepackage[switch]{lineno}
\usepackage{booktabs}
 \usepackage{multirow}
 \usepackage{pifont}
 \usepackage{bm}
 \usepackage{tabularx}
 \usepackage{adjustbox}
 \usepackage{makecell}
\usepackage{array}  
\usepackage{caption}
\usepackage{multirow}
\usepackage{colortbl}
\usepackage{tabularx}  
\usepackage{xcolor}   
\usepackage{diagbox}

\ifCLASSINFOpdf
\else
\fi

\def\ourmethod{Synthesis4AD}

\begin{document}

\title{Synthesis4AD: Synthetic Anomalies are All You Need for 3D Anomaly Detection}

\author{Yihan Sun\IEEEauthorrefmark{1},~\IEEEmembership{Student Member,~IEEE}, Yuqi~Cheng\IEEEauthorrefmark{1},~\IEEEmembership{Graduate Student Member,~IEEE},
Junjie Zu, Yuxiang Tan, \\
Guoyang Xie,~\IEEEmembership{Member,~IEEE}, Yucheng Wang, 
Yunkang~Cao\IEEEauthorrefmark{2},~\IEEEmembership{Member,~IEEE}, Weiming Shen\IEEEauthorrefmark{2},~\IEEEmembership{Fellow,~IEEE}
\thanks{*These authors contributed equally to this work.}
\thanks{$^\dagger$Co-corresponding authors.}
\thanks{
Manuscript received xxxx; revised xxxx; accepted xxxx. 
This work was supported by Fundamental Research Funds for the Central Universities (HUST: 2021GCRC058) and was part by the HPC Platform of Huazhong University of Science and Technology where the computation is completed. (\textit{Corresponding author: Yunkang Cao and Weiming Shen})

Yihan Sun, Yuqi Cheng, Junjie Zu, Yuxiang Tan and Weiming Shen are with the National Center of Technology Innovation for Intelligent Design and Numerical Control, Huazhong University of Science and Technology, Wuhan 430074, China (e-mail: yihansun@hust.edu.cn; yuqicheng@hust.edu.cn; jjzu@hust.edu.cn; yuxiangtan@hust.edu.cn;  wshen@ieee.org).

Yunkang Cao is with the School of Artificial Intelligence and Robotics, Hunan University, Changsha 410082, China (e-mail: caoyunkang@ieee.org).

Guoyang Xie is with the Department of Intelligent Manufacturing, Contemporary Amperex Technology Ltd., Ningde 352000, China (e-mail: guoyang.xie@ieee.org).

Yucheng Wang is with the Institute for Infocomm Research, A*STAR, Singapore 138632, and also with the School of Electrical and Electronic Engineering, Nanyang Technological University, Singapore 639798 (e-mail: yucheng003@e.ntu.edu.sg).

This work has been submitted to the IEEE for possible publication. Copyright may be transferred without notice, after which this version may no longer be accessible.
 
}}

\markboth{}{}

\maketitle

\IEEEpeerreviewmaketitle
\begin{abstract}

Industrial 3D anomaly detection performance is fundamentally constrained by the scarcity and long-tailed distribution of abnormal samples. To address this challenge, we propose \textit{Synthesis4AD}, an end-to-end paradigm that leverages large-scale, high-fidelity synthetic anomalies to learn more discriminative representations for 3D anomaly detection. At the core of Synthesis4AD is \textit{3D-DefectStudio}, a software platform built upon the controllable synthesis engine \textit{MPAS}, which injects geometrically realistic defects guided by higher-dimensional support primitives while simultaneously generating accurate point-wise anomaly masks. Furthermore, Synthesis4AD incorporates a multimodal large language model (MLLM) to interpret product design information and automatically translate it into executable anomaly synthesis instructions, enabling scalable and knowledge-driven anomalous data generation. To improve the robustness and generalization of the downstream detector on unstructured point clouds, Synthesis4AD further introduces a training pipeline based on spatial-distribution normalization and geometry-faithful data augmentations, which alleviates the sensitivity of Point Transformer architectures to absolute coordinates and improves feature learning under realistic data variations. Extensive experiments demonstrate state-of-the-art performance on Real3D-AD, MulSen-AD, and a real-world industrial parts dataset. The proposed synthesis method MPAS and the interactive system 3D-DefectStudio will be publicly released at \url{https://github.com/hustCYQ/Synthesis4AD}
.

\end{abstract}

\begin{IEEEkeywords}
3D Anomaly Detection; Point Cloud Anomaly Detection; Anomaly Synthesis; Point Transformer; 
Multimodal Large Language Model
\end{IEEEkeywords}
\section{Introduction}

Industrial visual anomaly detection has long been constrained by the limited availability of anomalous samples~\cite{zhang2025diffusionad, yao2025global, Cao2025Personalizing, chen2024progressive}, and this challenge is further intensified in 3D anomaly detection, where data acquisition is substantially more demanding. As a result, unsupervised anomaly detection (UAD) has become a prevalent setting, where models are trained exclusively on normal data and are expected to flag abnormalities at deployment~\cite{liu2024landmark, wang2025m3dm, cao2024bias, PCB}. However, real-world failures are inherently \emph{open-ended} and \emph{long-tailed}, which induces a pronounced train--test distribution shift: models learned on a narrow normal manifold must generalize to diverse, previously unseen structural defects~\cite{liu2024deep, Huang_VAD, jiang2022masked, cdo}. Consequently, learning only from normal samples often provides insufficient inductive bias to form representations that are reliably sensitive to subtle geometric/structural deviations.

\begin{figure}[t!]
\centering\includegraphics[width=0.9\linewidth]{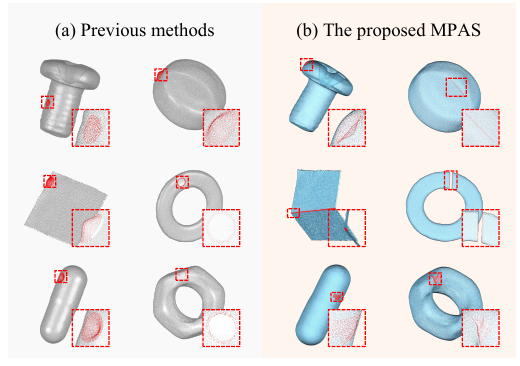}
\caption{Comparison between previous methods and our method in 3D anomaly synthesis. In (b), from top to bottom, the rows illustrate anomalies synthesized based on 1D, 2D, and 3D primitives, respectively.}
\label{fig:compare}
\end{figure}

To inject anomaly awareness, recent paradigms synthesize artificial defects during training. However, this strategy is inherently limited by a lack of \textit{physics-isomorphism}. As illustrated in the qualitative comparison in Fig.~\ref{fig:compare}(a), existing approaches predominantly rely on 1D primitives (e.g., singular points or thin curves) to anchor synthetic perturbations, as summarized in Table~1. This naive geometric tampering produces ``toy-level'' anomalies---such as homogeneous protrusions or regularized scratches---that are rigidly confined to highly localized regions. Because these formulations fail to model the spatially extended, complex topological distortions characteristic of real-world mechanical damage, their applicability to diverse industrial scenarios remains systematically bottlenecked.

\begin{table}[t]
\centering
\caption{Comparison of supported anomaly categories between current 3D anomaly synthesis methods and our MPAS method. MPAS enables diverse and realistic defect generation across multiple dimensions of primitives.} 
\label{table:comprision}
\fontsize{10}{14}\selectfont{
\resizebox{\linewidth}{!}{
\begin{tabular}{c|cc|cc|c}
\toprule[1.5pt]

\multirow{2}{*}{\diagbox[width=6em, height=3.5em, trim=l]{\textbf{Method}}{\textbf{Type}}} 
& \multicolumn{2}{c|}{\textbf{1D Primitives}} & \multicolumn{2}{c|}{\textbf{2D Primitives}} & \textbf{3D Primitives} \\

\cmidrule(lr){2-3} \cmidrule(lr){4-5} \cmidrule(lr){6-6}

 & Sphere & Scratch & Bent & Crack & Freeform Defect \\ 
\midrule

Group3D~\cite{Group3d}  & \ding{51}  & \ding{55}   & \ding{55}   & \ding{55}   & \ding{55}    \\ 
R3D-AD~\cite{r3d}  & \ding{51}  & \ding{55}   & \ding{55}   & \ding{55}   & \ding{55}    \\ 
GLFM~\cite{GLFM}  & \ding{51}  & \ding{55}   & \ding{55}   & \ding{55}   & \ding{55}    \\ 
MC4AD~\cite{MC4AD}  & \ding{51}  & \ding{51}   & \ding{55}   & \ding{55}   & \ding{55}    \\ 
PLANE~\cite{PLANE}  & \ding{51}  & \ding{51}   & \ding{55}   & \ding{55}   & \ding{55}    \\

\rowcolor{blue!8}
Ours  & \ding{51}  & \ding{51}   & \ding{51}   & \ding{51}   & \ding{51}    \\ 

\bottomrule[1.5pt]
\end{tabular}}}
\end{table}

To transcend these limitations, we propose \textbf{M}ulti-dimensional \textbf{P}rimitive-Guided \textbf{A}nomaly \textbf{S}ynthesis (MPAS), a framework that shifts the synthesis paradigm from heuristic local perturbations to physics-isomorphic geometric modeling. In contrast to prior approaches that rely on point- or curve-based supports, our method considers higher-dimensional geometric supports, such as surface regions or spatially bounded volumes, to guide anomaly generation. This design enables the synthesis of more complex defect patterns, including bending, structural damage, and free-form deformations, which are difficult to achieve with low-dimensional supports.
As shown in Fig.~1(b), our method produces anomalies with irregular boundaries, heterogeneous surface deformations, and coherent structural distortions that are well aligned with the underlying object geometry. This comprehensive synthesis capability provides more informative abnormal patterns and leads to more effective representation learning for 3D anomaly detection.

To bridge the gap between abstract engineering knowledge and concrete geometric generation, we propose \textit{Synthesis4AD}, an end-to-end procedural pipeline. At its core is \textit{3D-DefectStudio}, an interactive platform and programmatic Python API that enables scalable, high-fidelity defect injection and simultaneous ground-truth mask generation. Synthesis4AD fundamentally elevates this process by introducing a Multimodal Large Language Model (MLLM) as a cognitive-to-geometric translator. The MLLM parses abstract product design specifications, autonomously converting them into executable MPAS synthesis instructions. Furthermore, to stabilize learning across diverse geometric topologies, Synthesis4AD introduces a robust detector training pipeline utilizing spatial-distribution normalization (SDN) and geometry-faithful augmentations. During deployment, the learned representations are evaluated via prototype-based feature matching, yielding both point-wise localization maps and object-wise anomaly scores.

The main contributions of this work can be summarized as follows:
\begin{itemize}
\item We propose MPAS, a novel physics-isomorphic generation framework that mathematically models real-world defects across multi-dimensional primitives.
\item We develop \textit{Synthesis4AD} and its core engine \textit{3D-DefectStudio}, pioneering the first cognitive-to-geometric synthesis paradigm. By leveraging MLLMs, it translates abstract engineering priors into scalable, high-fidelity 3D defect generation.
\item We introduce a robust representation learning pipeline featuring Spatial-Distribution Normalization (SDN) and geometry-faithful augmentations. 
\item Extensive experiments demonstrate that our approach achieves state-of-the-art detection and localization performance on public benchmarks and exhibits exceptional generalization on real-world industrial scans.
\end{itemize}

The remainder of this paper is organized as follows. Section~\ref{sec:background} reviews related work on 3D anomaly detection and anomaly synthesis. Section~\ref{sec:method} introduces the proposed MPAS framework and details its design principles. Section~\ref{sec:system} introduces 3D-DefectStudio and the Synthesis4AD workflow, including the detector training pipeline. Experimental results on multiple benchmarks and real-world scenarios are reported in Section~\ref{sec:exp}. Finally, Section~\ref{sec:conclusion} concludes the paper and outlines directions for future work.

\section{Related work}\label{sec:background}

\subsection{Unsupervised 3D Anomaly Detection}
Unsupervised 3D anomaly detection ~\cite{MiniShift_Simple3D, MC3D} aims to distinguish abnormal samples from normal ones without access to defect annotations, where the core challenge lies in learning discriminative feature representations solely from normal data. Early methods predominantly relied on handcrafted geometric descriptors. For example, BTF~\cite{BTF} employs manually designed operators to characterize local geometric variations. However, such handcrafted features often lack sufficient expressiveness to capture complex defect patterns. 
Subsequent works have attempted to leverage deep point cloud encoders to improve feature representation. M3DM~\cite{M3DM} directly utilizes point features extracted from the pre-trained PointMAE~\cite{pointmae}.
Nevertheless, due to the limited availability of large-scale point cloud datasets, existing pre-trained models remain suboptimal in capturing fine-grained geometric anomalies, resulting in insufficient discriminative power for anomaly detection. To alleviate the scarcity of powerful 3D encoders, projection-based methods~\cite{CPMF,PCLIP,ISMP,PointAD} have been proposed to exploit the strong representation capabilities of 2D visual models. These approaches project 3D point clouds into multiple 2D views and apply pre-trained image encoders for feature extraction. While effective for surface-level inspection, such methods are inherently limited to 2.5D representations and struggle to faithfully model full 3D geometric structures.
Alternatively, some methods incorporate additional geometric priors to reduce feature ambiguity. Reg3D~\cite{read3d}, RegAD~\cite{RegAD}, and PointCore~\cite{patchcore} perform point cloud registration~\cite{gao2020complex, MVGR} prior to feature extraction, leveraging spatial correspondences to mitigate feature mismatching across samples. Although registration-aware strategies improve robustness to pose variations, they do not fundamentally address the limited anomaly sensitivity of the underlying feature representations.
Recognizing the critical role of representation learning, several works have attempted to explicitly train or fine-tune point cloud encoders to enhance their descriptive capacity. 3D-ST~\cite{STdes} introduces a self-supervised learning objective that encourages the encoder to reconstruct dense local descriptors. Similarly, Shape-Guided~\cite{shape} leverages neural implicit functions to supervise the training of a PointNet-based encoder. While these approaches enhance general geometric representation, they are primarily designed for shape understanding rather than anomaly awareness, and thus remain limited in distinguishing subtle abnormal structures from normal geometry.

\subsection{3D Anomaly Synthesis}
Anomaly synthesis has proven to be an effective strategy for enhancing feature discriminability in 2D anomaly detection~\cite{anomagic, jin2025dual, gui2024few, song2025defectfill}, and recent studies have extended this idea to the 3D domain. R3D-AD~\cite{r3d} synthesizes anomalies and employs a diffusion model to reconstruct abnormal inputs toward normality, thereby improving detection robustness. Group3D~\cite{Group3d} also utilizes synthetic anomalies, introducing Alignment Ranking Loss and Uniformity Ranking Loss to supervise the training of point cloud encoders. Further extending this direction, GLFM~\cite{GLFM} constructs point-level anomaly masks to formulate a segmentation task and fine-tunes a Point Transformer encoder, demonstrating that exposure to synthetic anomalies can significantly enhance anomaly sensitivity. In contrast, PO3AD~\cite{PO3AD} adopts a different strategy by directly supervising the learning of feature offsets using anomalous samples.
Despite these advances, the effectiveness of anomaly synthesis-based methods critically depends on the realism and diversity of the generated anomalies. Most existing approaches synthesize defects using simple geometric operations, such as spherical protrusions, holes, or scratches. These defect categories are limited in number and exhibit low intra-class diversity; for instance, protrusions are typically constrained to regular shapes such as circles or ellipses. However, real industrial products often exhibit complex defects induced by external forces, resulting in irregular deformations, structural bending, or free-form distortions that cannot be adequately described by existing synthesis strategies. Consequently, current methods fail to fully exploit the potential of anomaly synthesis for representation learning.

To address this gap, we conduct an in-depth analysis of defect patterns observed in real industrial scenarios and propose a high-fidelity anomaly synthesis framework that systematically models anomalies as geometry-guided deformations across different spatial extents. Furthermore, we release an open-source library and an interactive system to improve the accessibility and practicality of anomaly synthesis. Building upon this framework, we introduce Synthesis4AD as a representative example to demonstrate how large-scale, realistic synthetic anomalies can be effectively leveraged to train anomaly-aware point cloud encoders.

\section{MPAS Method}\label{sec:method}

\begin{figure*}[h!]
\centering\includegraphics[width=0.9\linewidth]{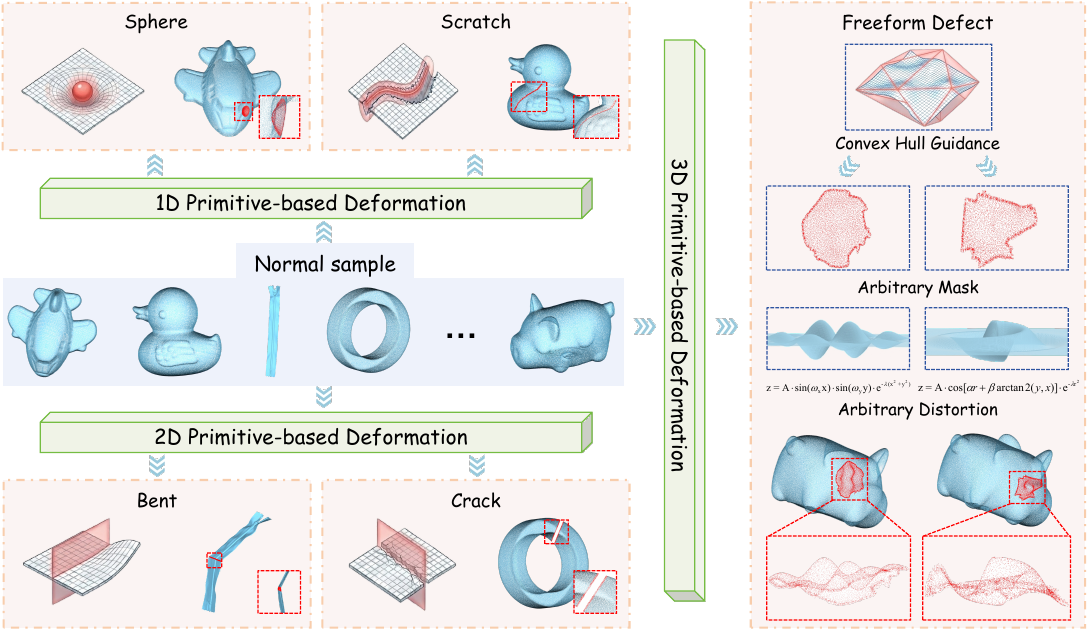}
\caption{MPAS framework leverages different dimensional primitives to automatically synthesize massive, realistic, and diverse 3D anomalous data.}
\vspace{-3mm}
\label{fig:Framework}
\end{figure*}

Existing 3D anomaly synthesis paradigms predominantly rely on single, regularized geometric perturbations, which inevitably yield morphologically homogeneous defects and often fail to preserve the physical plausibility consistent with real defect formation mechanisms. To address these limitations, we propose \textbf{M}ulti-dimensional \textbf{P}rimitive-Guided \textbf{A}nomaly \textbf{S}ynthesis (\textbf{MPAS}). As illustrated in Fig.~\ref{fig:Framework}, MPAS anchors deformation generation on supporting primitives and progressively expands the controllable deformation degrees of freedom across dimensions. Specifically, at the \textbf{1D level}, points or line segments are selected as supporting primitives to enable precise localization and fine-grained regulation of local deformations; at the \textbf{2D level}, planar supporting primitives are introduced to guide deformation propagation along stable surface patches, facilitating directional and continuous morphological variations; and in the \textbf{3D level}, the object convex hull serves as a global supporting primitives that imposes an explicit geometric shape prior, thereby providing a unified boundary for synthesizing more complex free-form distortions. Through this hierarchical primitive-driven design, MPAS integrates deformation onset, spatial extent, and magnitude into a unified parametric control framework, enabling the synthesis of complex anomalies with strong generative expressiveness while retaining structural interpretability and geometric credibility.

\subsection{1D Primitive-based Deformation}

MPAS exploits 1D primitives, denoted as MPAS-1D, to specifically synthesize point-like or line-like defects, including circular protrusions and depressions, holes, scratches, and grooves. MPAS-1D first selects multiple anchors to construct a deformation skeleton. This skeleton is then locally dilated into a neighborhood support to delineate the deformation region, which not only preserves the capability of generating discrete point anomalies but also enables continuous deformations with smoothly varying curvature. In particular, MPAS-1D can synthesize complex trajectories that traverse arbitrary surface topologies, thereby capturing the structural complexity and stochastic nature of real-world industrial defects. The overall pipeline is organized into the following stages:

\noindent\textbf{Multi-Anchor Sequence Selection:} 
To characterize the geometry of a 1D primitive, we employ an ordered sequence of $m$ anchor points, denoted as $\mathcal{A} = \{\boldsymbol{a}_1, \boldsymbol{a}_2, \dots, \boldsymbol{a}_m\} \subset \boldsymbol{P}$, where $m \ge 1$. This sequence forms the skeleton. MPAS-1D synthesizes either discrete point anomalies ($m=1$) or continuous, smooth surface deformations ($m>1$), depending on the number of anchor points.

\noindent\textbf{Composite Geodesic Path Construction:} 
To articulate consecutive anchors with a dense, geometry-consistent path on the point cloud surface, we build an undirected graph $G = (\{\boldsymbol{p}_i\}, \{\boldsymbol{e}_{ij}\})$ by defining edges $\boldsymbol{e}_{ij} = \langle \boldsymbol{p}_i, \boldsymbol{p}_j \rangle$, where $\boldsymbol{p}_j$ is among the $k$ nearest neighbors of $\boldsymbol{p}_i \in \boldsymbol{P}$. For linear primitives ($m \ge 2$), Dijkstra’s algorithm~\cite{Dijkstra} is employed to compute the shortest geodesic path $\pi^*_k$ between each pair of consecutive anchors $(\boldsymbol{a}_k, \boldsymbol{a}_{k+1})$. The final 1D support region $\boldsymbol{\Gamma}$ is defined as the union of these sub-paths (noting that for $m=1$, $\boldsymbol{\Gamma}$ degenerates to the single anchor point):
\begin{equation}
    \pi^*_k = \mathop{\arg\min}\limits_{\pi \in \Pi(\boldsymbol{a}_k, \boldsymbol{a}_{k+1})} \sum_{\boldsymbol{e}_{ij} \in \pi} w(\boldsymbol{e}_{ij})
\end{equation}
\begin{equation}
    \boldsymbol{\Gamma} = \bigcup_{k=1}^{m-1} \left\{ \boldsymbol{p}_j \in \boldsymbol{P} \mid \exists \boldsymbol{e}_{ij} \in \pi^*_k \text{ or } \boldsymbol{p}_j \in \{\boldsymbol{a}_k, \boldsymbol{a}_{k+1}\} \right\}
\end{equation}
where $\Pi(\boldsymbol{a}_k, \boldsymbol{a}_{k+1})$ denotes the set of all possible paths connecting $\boldsymbol{a}_k$ and $\boldsymbol{a}_{k+1}$, and $w(\boldsymbol{e}_{ij})$ represents the Euclidean distance between connected nodes. This formulation allows the synthesized anomaly to traverse complex geometries with arbitrary turning angles, significantly enriching the morphological variety.

\noindent\textbf{Region Expansion and Distortion:} 
Taking the established geodesic region $\boldsymbol{\Gamma}$ as the central axis, the deformation/mask region $\boldsymbol{M}_r$ is constructed by expanding outward from $\boldsymbol{\Gamma}$ with radius $r$, thereby including all points in $\boldsymbol{P}$ located within this neighborhood:

\begin{equation}
    \boldsymbol{M}_r = \left\{ \boldsymbol{p}_j \in \boldsymbol{P} \mid d_j < r \right\}, \quad d_j = \min_{\boldsymbol{p}_v \in \boldsymbol{\Gamma}} \|\boldsymbol{p}_j - \boldsymbol{p}_v\|
\end{equation}
The distortion direction $\bar{\boldsymbol{n}}_{\text{avg}}$ is determined by normalizing the average normal of points within $\boldsymbol{M}_r$:
\begin{equation}
    \bar{\boldsymbol{n}}_{\text{avg}} = \frac{\bar{\boldsymbol{n}}}{\|\bar{\boldsymbol{n}}\|_2}, \quad \bar{\boldsymbol{n}} = \frac{1}{|\boldsymbol{M}_r|} \sum_{\boldsymbol{p}_j \in \boldsymbol{M}_r} \boldsymbol{n}_j
\end{equation}
We then synthesize the anomaly by displacing each point along $\bar{\boldsymbol{n}}_{\text{avg}}$ according to its distance-dependent weight. Specifically, a linearly decaying displacement field is applied so that points closer to the central skeleton undergo larger deformation, while those near the boundary are progressively attenuated, thereby ensuring a smooth geometric transition:

\begin{equation}
    \boldsymbol{p}_j^{\prime} = \boldsymbol{p}_j + dir \cdot \bar{\boldsymbol{n}}_{\text{avg}} \cdot \left( 1 - \frac{d_j}{d_{\text{max}}} \right) \cdot d
\end{equation}
where $dir \in \{1, -1\}$ controls the deformation polarity, corresponding to protrusive and depressive anomalies, respectively; $d$ denotes the peak deformation magnitude; and $d_{\text{max}} = \max_{\boldsymbol{p}_k \in \boldsymbol{M}_r} d_k$ is the maximum distance within the mask region used for normalization.

\subsection{2D Primitive-based Deformation}

Compared with MPAS-1D,  MPAS-2D provides a natural extension for modeling more global structural anomalies under planar geometric constraints. Given a primitive plane, MPAS-2D first extracts its intersection band with the object point cloud, and then uses this band either as a fracture region for crack synthesis or as a hinge region for bending synthesis. The details are described below:

\noindent\textbf{Primitive Plane Instantiation.}
We define a 2D primitive as a plane:
\begin{equation}
\Pi(\mathbf{n}, \mathbf{c}) = \left\{ \mathbf{x} \in \mathbb{R}^3 \mid \mathbf{n}^{\top}(\mathbf{x}-\mathbf{c}) = 0 \right\},
\end{equation}
where $\mathbf{n}$ is the unit normal vector and $\mathbf{c}$ is a point on the plane.
For each point $\mathbf{p}_i \in \mathbf{P}$, we compute its signed distance to the plane:
\begin{equation}
s_i =  \mathbf{n}^{\top}(\mathbf{p}_i - \mathbf{c}).
\end{equation}

\noindent\textbf{Plane-Intersection Band Extraction.}
Since the intersection between a plane and a discrete point cloud is observed as a thin band rather than an ideal analytic curve, we extract a narrow intersection band around the plane:
\begin{equation}
\mathcal{B}_{\delta} = \left\{ \mathbf{p}_i \in \mathbf{P} \mid |s_i| < \frac{\delta}{2} \right\},
\end{equation}
where $\delta$ controls the thickness of the extracted band. This band serves as the common geometric basis for both bending and cracking generation.

\noindent\textbf{Bending Generation.}
For bending synthesis, the intersection band $\mathcal{B}_{\delta}$ is first used to estimate a geometrically meaningful hinge axis. Specifically, we compute the centroid of the band
\begin{equation}
\mathbf{x}_0 = \frac{1}{|\mathcal{B}_{\delta}|} \sum_{\mathbf{p}_i \in \mathcal{B}_{\delta}} \mathbf{p}_i,
\end{equation}
and fit a 3D line to $\mathcal{B}_{\delta}$ via PCA. Let $\mathbf{h}$ denote the dominant principal direction of the covariance matrix of $\mathcal{B}_{\delta}$; the hinge axis is then defined as:
\begin{equation}
\ell(t) = \mathbf{x}_0 + t \mathbf{h}, \quad t \in \mathbb{R}.
\end{equation}

We then partition the point cloud according to the signed distance $s_i$ and define a continuous angular weight
\begin{equation}
\alpha_i =
\begin{cases}
0, & s_i \le -\frac{\delta}{2}, \\[4pt]
\dfrac{s_i + \delta/2}{\delta}, & -\frac{\delta}{2} < s_i < \frac{\delta}{2}, \\[8pt]
1, & s_i \ge \frac{\delta}{2},
\end{cases}
\end{equation}
so that points on one side of the plane remain unchanged, points on the other side undergo the full bending rotation, and points inside the intersection band transition smoothly between the two states. The deformed point is given by:
\begin{equation}
\mathbf{p}_i' = \mathbf{x}_0 + \mathbf{R}(\alpha_i \theta, \mathbf{h}) \left( \mathbf{p}_i - \mathbf{x}_0 \right),
\end{equation}
where $\mathbf{R}(\alpha_i \theta, \mathbf{h})$ denotes the rotation matrix around axis $\mathbf{h}$ with angle $\alpha_i \theta$, and $\theta$ is the maximum bending angle.

The corresponding anomaly mask is defined as:
\begin{equation}
\mathcal{M}_{\text{bend}} = \left\{ \mathbf{p}_i \in \mathbf{P} \mid \alpha_i > 0 \right\}.
\end{equation}

\noindent\textbf{Cracking Generation.}
For crack synthesis, the same intersection band is treated as the fracture support. To simulate irregular fracture boundaries, we perturb the signed distance with a stochastic offset:
\begin{equation}
\tilde{s}_i = s_i + \eta_i, \quad \eta_i \sim \mathcal{N}(0, \sigma^2).
\end{equation}
The points to be removed are then defined as:
\begin{equation}
\mathcal{S}_{\text{remove}} = \left\{ \mathbf{p}_i \in \mathbf{P} \mid |\tilde{s}_i| < \frac{\tau}{2} \right\},
\end{equation}
where $\tau$ denotes the crack width. The cracked point cloud is obtained by:
\begin{equation}
\mathbf{P}' = \mathbf{P} \setminus \mathcal{S}_{\text{remove}}.
\end{equation}

To provide point-wise supervision on the retained point cloud, we define the crack mask as the boundary band adjacent to the removed region:
\begin{equation}
\mathcal{M}_{\text{crack}} =
\left\{
\mathbf{p}_i \in \mathbf{P}' \mid
\frac{\tau}{2} \le |\tilde{s}_i| < \frac{\tau}{2} + r_c
\right\},
\end{equation}
where $r_c$ controls the thickness of the labeled crack boundary region.

\subsection{3D Primitive-based Deformation}

While MPAS-1D and MPAS-2D effectively capture localized line-like defects and plane-induced structural anomalies, many real-world industrial defects exhibit spatially extended, free-form geometric deviations that cannot be adequately described by low-dimensional supports. To model such irregularities, we further introduce MPAS-3D, a high-flexibility deformation module that employs a 3D support primitive to define an adaptive surface patch and then applies locally parameterized free-form distortion followed by geometric regularization. Unlike fixed analytic primitives, this formulation decouples the support region from a predefined defect template, thereby enabling the synthesis of diverse yet geometrically coherent anomalies with higher morphological complexity and stronger physical plausibility. The pipeline consists of three stages: convex hull-guided mask generation, local parametric surface distortion, and local surface smoothing.

\noindent\textbf{Convex Hull-Guided Mask Generation.}
To obtain a spatially adaptive support region with irregular boundaries, we instantiate the 3D primitive using a stochastic anchor set:
\begin{equation}
\mathcal{A} = \{\mathbf{a}_1, \dots, \mathbf{a}_m\} \subset \mathbf{P},
\end{equation}
where the anchor points are sampled from the input point cloud and serve as control vertices for defining the deformation extent. Based on these anchors, we construct a convex hull $\mathcal{H}$~\cite{barber1996quickhull}, defined as the set of all convex combinations of $\mathcal{A}$:
\begin{equation}
\mathcal{H} =
\left\{
\sum_{k=1}^{m} \alpha_k \mathbf{a}_k
\;\middle|\;
\sum_{k=1}^{m} \alpha_k = 1,\;
\alpha_k \ge 0
\right\}.
\end{equation}

Rather than treating $\mathcal{H}$ as a solid volumetric region, we use its boundary surface $\partial \mathcal{H}$ as a geometric support to localize surface-level defects. The anomaly mask is then defined as the set of points lying within a narrow neighborhood of this boundary:
\begin{equation}
\mathcal{M}_{\mathcal{H}} =
\left\{
\mathbf{p}_i \in \mathbf{P}
\;\middle|\;
\min_{\mathbf{q} \in \partial \mathcal{H}} \|\mathbf{p}_i - \mathbf{q}\|_2 < \epsilon
\right\},
\end{equation}
where $\epsilon$ is a proximity threshold controlling the thickness of the selected surface patch. In this way, the support region is determined by the spatial arrangement of the anchors rather than by a hand-crafted geometric template, allowing the mask to adapt naturally to non-planar and spatially irregular local structures.

\noindent\textbf{Local Parametric Surface Distortion.}
Once the support mask $\mathcal{M}_{\mathcal{H}}$ is established, we synthesize the defect by defining a free-form displacement field over the selected surface patch. To this end, we first construct a local tangent coordinate system via PCA~\cite{PCA} over $\mathcal{M}_{\mathcal{H}}$. Let $\mathbf{c}$ denote the centroid of the masked subset, and let $\{\mathbf{u}, \mathbf{v}\}$ denote the two dominant principal directions spanning the local tangent plane. Each point $\mathbf{p}_i \in \mathcal{M}_{\mathcal{H}}$ is then projected onto this local frame to obtain 2D coordinates:
\begin{equation}
u_i = (\mathbf{p}_i - \mathbf{c})^\top \mathbf{u},
\qquad
v_i = (\mathbf{p}_i - \mathbf{c})^\top \mathbf{v}.
\end{equation}

On this local parameter domain, we define a scalar height field $h: \mathbb{R}^2 \rightarrow \mathbb{R}$ to control the deformation profile. To achieve high expressiveness while retaining smoothness, $h(u,v)$ is modeled as a superposition of Gaussian basis functions:
\begin{equation}
h(u_i, v_i) = \sum_{k=1}^{K}
A_k \exp \left( -\frac{(u_i - \mu_{u,k})^2 + (v_i - \mu_{v,k})^2}{2\sigma_k^2}
\right),
\end{equation}
where $K$ is the number of kernels, and $A_k$, $(\mu_{u,k}, \mu_{v,k})$, and $\sigma_k$ denote the amplitude, center, and spatial spread of the $k$-th component, respectively. By varying the number, sign, scale, and spatial arrangement of these kernels, the proposed formulation can generate a broad family of multi-modal surface profiles, ranging from simple dents and bulges to compound free-form distortions.

To preserve the local surface geometry, the displacement is applied along the point normal direction $\mathbf{n}_i$:
\begin{equation}
\tilde{\mathbf{p}}_i = \mathbf{p}_i + h(u_i, v_i)\mathbf{n}_i.
\end{equation}
This normal-aligned update ensures that the deformation follows the local surface manifold instead of introducing arbitrary Cartesian offsets that may break geometric consistency.

\noindent\textbf{Local Surface Smoothing.}
Directly applying a highly flexible parametric deformation may produce undesired high-frequency artifacts, especially near the boundary of the selected mask. To improve geometric continuity and better mimic naturally occurring material deformation, we perform a local smoothing step on the deformed subset. Specifically, we construct a $k$-nearest-neighbor graph over the intermediate points $\{\tilde{\mathbf{p}}_i\}$ and denote the neighbor set of $\tilde{\mathbf{p}}_i$ by $\mathcal{N}_i$. The final point position is updated by distance-weighted averaging:
\begin{equation}
\mathbf{p}'_i
= (1-\lambda)\tilde{\mathbf{p}}_i + \lambda \sum_{j \in \mathcal{N}_i}
w_{ij}\tilde{\mathbf{p}}_j,
\qquad w_{ij} = \frac{d_{ij}^{-1}}{\sum_{l \in \mathcal{N}_i} d_{il}^{-1}},
\end{equation}
where $\lambda \in [0,1]$ controls the smoothing strength and
\begin{equation}
d_{ij} = \|\tilde{\mathbf{p}}_i - \tilde{\mathbf{p}}_j\|_2.
\end{equation}
This regularization suppresses spurious local oscillations while preserving the overall defect morphology, thus allowing the synthesized anomaly to blend more naturally with the surrounding surface.

In summary, MPAS-3D provides the highest representational flexibility within the MPAS hierarchy by enabling adaptive support selection and free-form geometric deformation on arbitrarily shaped surface patches. More importantly, it can be viewed as a generalized extension of the lower-dimensional primitives: when the support region degenerates to a narrow curve-like neighborhood or a plane-constrained band, the resulting deformation behavior approaches that of the 1D and 2D cases. This hierarchical design substantially enriches the diversity of synthetic anomalies and improves the coverage of complex defect distributions encountered in real industrial inspection scenarios.

\section{Synthesis4AD System}
\label{sec:system}

\begin{figure*}[h!]
\centering
\includegraphics[width=\linewidth]{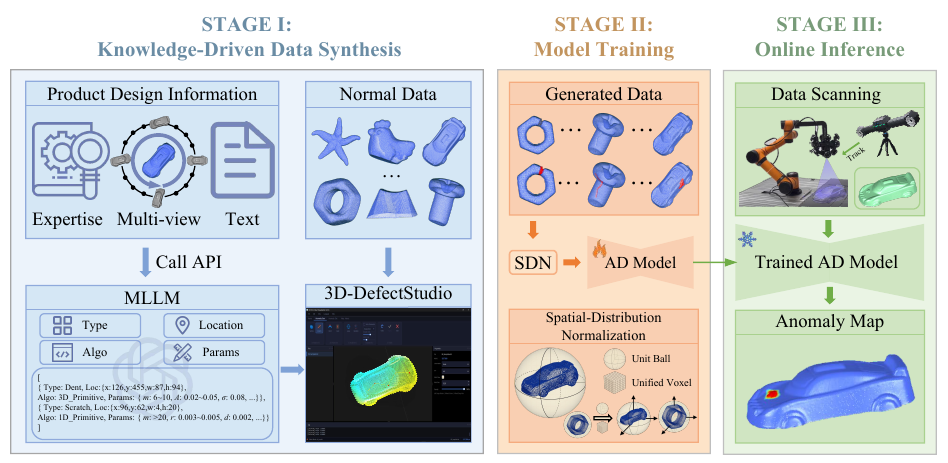}
\caption{Overview of the proposed \textbf{Synthesis4AD} system. Stage I parses product-side knowledge, including expert priors, multi-view cues, and textual specifications, into executable synthesis instructions via an MLLM, and drives 3D-DefectStudio to inject controllable anomalies into large-scale normal 3D assets. Stage II trains the anomaly detector using the generated anomalous samples and their ground-truth masks. Stage III deploys the trained model for prototype-based online inference, producing both point-wise anomaly maps and object-wise anomaly scores from scanned test data.}
\label{fig:System}
\end{figure*}

To move the proposed MPAS pipeline from an algorithmic design to a practically deployable industrial workflow, we further develop an end-to-end system, termed \textbf{Synthesis4AD}, which unifies anomaly synthesis, detector training, and online inference within a single framework. Specifically, MPAS is encapsulated as a reusable Python library and an interactive software platform, \textit{3D-DefectStudio}, while an MLLM is introduced to transform multimodal product knowledge into structured synthesis instructions that can be directly executed by the MPAS backend. In this manner, Synthesis4AD enables the automatic generation of diverse, controllable, and labeled anomalous 3D samples from massive normal assets, thereby providing scalable supervision for downstream 3D anomaly detection. As illustrated in Fig.~\ref{fig:System}, Synthesis4AD is organized as a three-stage and tightly coupled pipeline that progressively transforms product-side knowledge into executable anomaly synthesis, then into effective supervisory signals for detector learning, and finally into reliable defect inference on real scanned data.

\subsection{Stage I: Knowledge-Driven Data Synthesis}

The first stage converts product knowledge into executable anomaly synthesis programs. Its inputs include: (i) design-side multimodal knowledge, and (ii) large collections of normal point clouds. The design-side knowledge is represented as:
\begin{equation}
\mathcal{Z}=\left\{\mathcal{I}_{\text{mv}},\mathcal{T}_{\text{spec}},\mathcal{E}_{\text{prior}}\right\},
\end{equation}
where $\mathcal{I}_{\text{mv}}$ denotes multi-view visual observations of the target asset (e.g., rendered views or CAD snapshots), $\mathcal{T}_{\text{spec}}$ denotes textual specifications such as structural descriptions, tolerance constraints, and surface requirements, and $\mathcal{E}_{\text{prior}}$ denotes expert knowledge about plausible defect modes and critical regions. These complementary cues jointly inform \emph{what} kind of anomaly should be synthesized, \emph{where} it should be placed, and \emph{how} strong the geometric perturbation should be.

Given $\mathcal{Z}$, the MLLM acts as a semantic parser to generate a machine-executable instruction tuple:
\begin{equation}
\mathbf{c}=\Phi_{\psi}(\mathcal{Z}) = \left\{u,\mathcal{R},\mathcal{A},\Theta\right\},
\end{equation}
where $\Phi_{\psi}$ is the MLLM-based parsing function, $u$ specifies the anomaly type, $\mathcal{R}$ denotes the suggested anomaly region, $\mathcal{A}$ specifies the algorithm interface to be invoked in 3D-DefectStudio (e.g., the corresponding MPAS-1D/2D/3D operator), and $\Theta$ denotes the synthesis parameter set, which may include operator-dependent geometric attributes such as length, width, depth, deformation magnitude, and other control variables. This structured representation serves as an intermediate program layer that bridges semantic understanding and deterministic geometric synthesis.

To ensure the generated instruction is physically executable on a target point cloud $X$, Synthesis4AD introduces a validation module $\mathcal{V}(\cdot, X)$. This module performs schema checking to enforce format consistency and geometry-aware grounding to validate topological feasibility. Instructions that fail validation are replaced by predefined rule-based templates from the operator library. The final executable instruction $\mathbf{c}^{\ast}$ is formulated as:
\begin{equation}
\mathbf{c}^{\ast}=
\begin{cases}
\mathbf{c}, & \text{if } \mathcal{V}(\mathbf{c}, X) \text{ is True},\\
\mathbf{c}_{\text{rule}}, & \text{otherwise},
\end{cases}
\end{equation}
where $\mathbf{c}_{\text{rule}}$ denotes a valid fallback instruction. This mechanism explicitly prevents physically implausible synthesis such as invalid placements or excessive deformations.

Based on the validated instruction, 3D-DefectStudio calls the corresponding MPAS interface to inject anomalies into the normal sample and simultaneously produce the associated ground-truth annotation:
\begin{equation}
\left(X^{\text{anom}},\,M^{\text{gt}}\right)=\mathcal{G}_{\text{MPAS}}(X,\mathbf{c}^{\ast}),
\end{equation}
where $\mathcal{G}_{\text{MPAS}}(\cdot)$ denotes the deterministic anomaly generation operator implemented in the backend, $X^{\text{anom}}$ is the synthesized anomalous point cloud, and $M^{\text{gt}}$ is its point-wise defect mask. Through this process, Stage I converts large-scale normal 3D assets into paired anomaly--annotation data suitable for downstream detector learning.

\subsection{Stage II: Model Training}

The synthesized anomaly--mask pairs generated in Stage I are subsequently used to train the downstream anomaly detector. By exposing the model to diverse anomaly types, spatial distributions, and geometric severities, Synthesis4AD provides effective supervision for learning discriminative 3D feature representations that generalize beyond a fixed product category.

The detector adopts a Point Transformer~\cite{pointmae} architecture for hierarchical point feature encoding, followed by a lightweight MLP-based segmentation head for dense point-wise supervision during training. Following the training protocol of GLFM~\cite{GLFM}, each synthesized anomalous point cloud is fed into the encoder to produce contextualized point-wise representations, and the segmentation head predicts a dense binary map that is optimized against the corresponding synthetic anomaly mask. In this way, the detector directly exploits large-scale, high-fidelity synthetic anomalies to learn task-aligned local geometric cues. It is worth noting that the segmentation head is introduced to inject dense supervisory signals during training, whereas the learned encoder is later reused in Stage III for feature-space prototype matching.

Despite the availability of large-scale synthetic supervision, directly training Point Transformer backbones on raw point clouds remains non-trivial. First, point clouds are inherently unstructured, and industrial categories often exhibit substantial variations in object scale and sampling density, which can lead to unstable optimization and poor cross-category generalization. In addition, Point Transformer-style positional encodings are sensitive to absolute coordinates, making the learned features brittle, especially when the object pose changes at test time. To alleviate these issues, we introduce Spatial-Distribution Normalization (SDN) and a set of training-time data augmentations to regularize point statistics and reduce over-reliance on absolute coordinates.

\noindent\textbf{Spatial-Distribution Normalization (SDN).}
Given an input point cloud $P=\{\mathbf{x}_i\}_{i=1}^{N}$ with $\mathbf{x}_i\in\mathbb{R}^3$, SDN first maps each category into a canonical unit-ball space and then performs voxel downsampling with a unified voxel resolution. Specifically, for each category $c$, we compute its category-level bounding sphere characterized by a center $\mathbf{o}_c$ and a radius $r_c$. Each point is then normalized as:
\begin{equation}
\tilde{\mathbf{x}}_i = \frac{\mathbf{x}_i-\mathbf{o}_c}{r_c}, \qquad \tilde{P}=\{\tilde{\mathbf{x}}_i\}_{i=1}^{N},
\label{eq:sdn_norm}
\end{equation}
such that $\tilde{P}$ lies within the unit ball, i.e., $\|\tilde{\mathbf{x}}_i\|_2\leq 1$. We subsequently apply voxel downsampling on $\tilde{P}$ with a fixed voxel size $v_0$ shared across all categories, yielding a subsampled point set $P'=\{\mathbf{x}'_j\}_{j=1}^{N'}$ with $N'\ll N$. In this way, SDN enforces a consistent effective geometric granularity across categories and reduces the scale-induced bias in coordinate statistics, which stabilizes training and improves cross-category generalization.

\noindent\textbf{Data Augmentations.}
After SDN, we further apply a set of augmentations to improve robustness to pose changes and sensing artifacts commonly encountered in real acquisition. Concretely, for the normalized point set $P'$, we utilize three standard transformations targeting specific geometric variations: (1) \textbf{Random Rotation} to mitigate sensitivity to global object pose; (2) \textbf{Noise Perturbation} to simulate measurement uncertainty; and (3) \textbf{Point Dropout} to mimic missing observations from self-occlusions or scanning limitations. In practice, these transformations are jointly applied during training. They encourage the encoder to learn local geometric representations that are less sensitive to global pose, while remaining robust to sensor noise and partial observations.

\subsection{Stage III: Online Inference}

After Stage II, the encoder has been trained to produce discriminative point-wise embeddings under dense synthetic supervision. To convert these learned representations into a practical detection mechanism, Synthesis4AD performs inference through a prototype-based matching scheme in the learned feature space.

Before deployment, the trained encoder is first applied to normal training point clouds to extract point-wise features and construct a normal prototype set:
\begin{equation}
\mathcal{Q}=\{\mathbf{q}_k\}_{k=1}^{K},
\end{equation}
which summarizes the distribution of normal local geometry in the embedding space. This prototype set serves as the reference model of normality for subsequent anomaly scoring.

During online inference, a scanned test point cloud $X_{test}$ is first processed using the same normalization pipeline as in Stage II, ensuring consistent geometric statistics between training and deployment. The normalized test sample is then passed through the trained encoder to obtain point-wise features:
\begin{equation}
\mathcal{F}(X_{test})=\{\mathbf{f}_i\}_{i=1}^{N}.
\end{equation}
For each point feature $\mathbf{f}_i$, its anomaly score is computed according to its deviation from the normal prototype set:
\begin{equation}
s_i=\min_{1\leq k\leq K} d(\mathbf{f}_i,\mathbf{q}_k),
\end{equation}
where $d(\cdot,\cdot)$ denotes the feature-space distance metric. Intuitively, points that can be well explained by the normal prototypes receive low scores, whereas points that deviate significantly from the normal feature distribution are assigned higher anomaly scores.

The point-wise scores $\{s_i\}_{i=1}^{N}$ form a dense anomaly map for localized defect indication and visual assessment. To further support instance-level quality inspection, these point-wise responses are aggregated into an object-wise anomaly score:
\begin{equation}
S(X_{test})=\operatorname{Agg}\!\left(\{s_i\}_{i=1}^{N}\right),
\end{equation}
where $\operatorname{Agg}(\cdot)$ computes the mean of the top-$K$ highest point-wise scores. In this way, Stage III simultaneously provides fine-grained defect localization and holistic sample-level abnormality estimation. 

By tightly coupling knowledge-guided synthesis, supervised feature learning, and prototype-based deployment, Synthesis4AD forms a scalable and practical pipeline for 3D anomaly detection in complex industrial scenarios.
\section{Experiment}\label{sec:exp}

To comprehensively evaluate Synthesis4AD, we conduct experiments from four complementary perspectives: (i) assessing the realism of the synthesized anomalies, (ii) benchmarking \ourmethod{} on public 3D anomaly detection datasets, (iii) performing systematic ablations to quantify the impact of key design choices, and (iv) validating practical applicability through real-world experiments.

\subsection{Evaluation of Synthetic Anomaly Realism}

\begin{figure*}[t!]
\centering\includegraphics[width=0.9\linewidth]{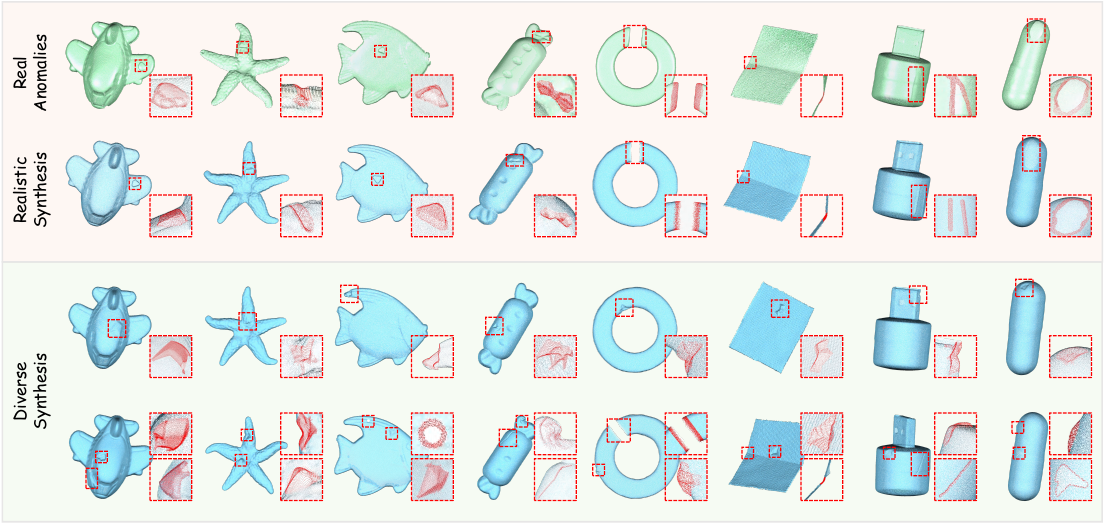}
\caption{\textbf{Visualization of anomalies.} From top to bottom: real anomalies, Synthesized anomalies by MPAS with the same types, and two rows of more diverse compound anomalies synthesized by MPAS. Red insets highlight defect regions for detailed comparison.}
\label{fig:Vis_anomaly}
\end{figure*}

\begin{figure}[t!]
\centering\includegraphics[width=\linewidth]{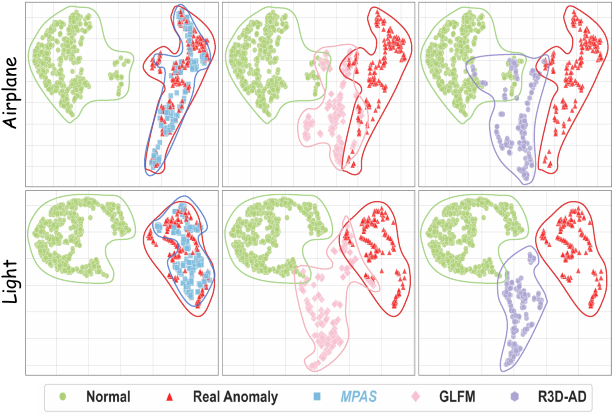}
\caption{t-SNE visualization of feature distributions. Normal samples and real anomalies are compared with synthetic anomalies generated by MPAS, GLFM, and R3D-AD.}
\label{fig:Vis_anomaly_feature}
\end{figure}

Anomaly realism is a prerequisite for using synthesized defects as effective supervision, since unrealistic perturbations tend to introduce spurious cues and yield biased representations. Fig.~\ref{fig:Vis_anomaly} therefore visualizes our synthesis results against real defects. Real anomalies are shown as references, and MPAS is then applied to synthesize the same defect types under comparable geometric conditions; the magnified views reveal that the resulting distortions and boundary profiles closely follow those observed in real defects, indicating that MPAS can faithfully reproduce realistic defect morphologies at the geometric level. Importantly, MPAS is not limited to replicating existing patterns: by enabling compositional synthesis, it further produces more complex and heterogeneous anomalies that combine multiple deformation modes, as illustrated by the additional rows. These compound defects substantially enrich the anomaly space and compensate for the limited coverage and diversity of real defect collections, thereby providing more challenging training signals for downstream anomaly detection.

The same conclusion is supported when the comparison is moved from geometry to representation space. We extract features of normal samples, real anomalies, and anomalies generated by different synthesis strategies using the Synthesis4AD model trained in this work, and project them using t-SNE for distributional inspection. As shown in Fig.~\ref{fig:Vis_anomaly_feature}, generated anomalies consistently lie closer to the real-anomaly clusters and exhibit stronger overlap with them than competing synthesis methods, which either drift toward the normal manifold or form separated clusters. This feature-level alignment suggests that MPAS not only produces visually plausible defects, but also preserves the statistical characteristics of real anomalies that matter for representation learning.

\subsection{Performance on Public Benchmarks}

\subsubsection{Experimental Settings}

We evaluate the proposed method on two public 3D anomaly detection benchmarks, Real3D-AD~\cite{read3d} and MulSen-AD~\cite{MulSen-AD}. Both datasets provide normal samples for training and contain anomalous instances for evaluation, together with point-wise anomaly annotations that enable fine-grained localization assessment. To quantify performance at different granularities, we report object-wise AUROC (O-ROC) and point-wise AUROC (P-ROC). O-ROC measures how well the model separates normal and anomalous objects using a single anomaly score per sample, whereas P-ROC evaluates the quality of localization by comparing predicted point-wise anomaly scores against the ground-truth masks across all points.

\noindent\textbf{Implementation Details}
Synthesis4AD instantiates the MLLM component with Gemini 3~\cite{team2023gemini}. Guided by the category metadata provided by each benchmark, the MLLM generates structured synthesis instructions and invokes the 3D-DefectStudio APIs to construct 5,000 and 20,000 anomalous point clouds for Real3D-AD and MulSenAD, respectively. The feature extractor is then trained on these synthesized samples for 30k iterations with a batch size of 4 and a learning rate of 1e-5. All experiments are conducted on an NVIDIA RTX 5880 Ada (48GB) GPU. 

\noindent\textbf{Comparison Studies}
For a comprehensive comparison, we benchmark against representative 3D anomaly detection methods including the widely used memory-bank baseline PatchCore~\cite{patchcore}, instantiated with different feature representations: using handcrafted FPFH~\cite{FPFH} descriptors (PatchCore-FP), combining FPFH with additional raw geometric cues (PatchCore-FP-R), and using features extracted from a PointMAE~\cite{pointmae} backbone (PointCore-PM). We further compare with multi-view projection-based methods, including CPMF~\cite{CPMF} and ISMP~\cite{ISMP}, as well as reconstruction-based approaches such as IMRNet~\cite{shape_anomaly} and MC3D-AD~\cite{MC3D}. In addition, we consider methods that exploit geometric alignment cues, including the registration-enhanced Reg3D-AD~\cite{read3d} and GroupAD~\cite{Group3d}. Finally, we also include R3D-AD~\cite{r3d}, GLFM~\cite{GLFM}, and PO3AD~\cite{PO3AD} to reflect the line of work that improves feature discriminability through synthesized anomalies.

\subsubsection{Experimental Results}

\begin{table*}[t]
\centering
\caption{\textbf{Quantitative Results on Real3D-AD.} The results are presented in O-ROC\%/P-ROC\%. The best performance is in \textbf{bold}, and the second best is \underline{underlined}.}
\label{table:real}
\fontsize{10}{14}\selectfont{
\resizebox{\linewidth}{!}{
\begin{tabular}{c|cc
>{\columncolor{blue!8}}c
ccccccc|
>{\columncolor{blue!8}}c}
\toprule[1.5pt]

Method~$\rightarrow$   & CPMF & R3D-AD & R3D-AD & Reg3D-AD & Group3AD & IMRNet & ISMP & PO3AD & MC3D-AD & GLFM & \textbf{\ourmethod} \\
Category~$\downarrow$  & PR'24 & ECCV'24 & +MPAS & NeurIPS'23 & ACM MM'24 & CVPR'24 & AAAI'25 & CVPR'25 & IJCAI'25 & TASE'25 & \textbf{Ours} \\ \midrule
Airplane & 63.2/61.8 & 76.5/61.4 & 81.3/61.4 & 71.6/63.1 & 74.4/63.6 & 76.2/- & \textbf{85.8}/\underline{75.3} & 80.4/- & \underline{85.0}/62.8 & 54.6/74.3 & 62.1/\textbf{83.6} \\
Car      & 51.8/83.6 & 68.2/62.0 & 74.4/62.0 & 69.7/71.8 & 72.8/74.5 & 71.1/- & 73.1/83.6 & 65.4/- & 74.9/81.9 & \underline{84.2}/\underline{88.2} & \textbf{96.3}/\textbf{95.7} \\
Candybar & 71.8/73.4 & 66.4/61.2 & 72.0/61.2 & 82.7/72.4 & \underline{84.7}/73.8 & 75.5/- & \textbf{85.2}/\underline{90.7} & 78.5/- & 83.0/\textbf{91.0} & 71.5/79.7 & 79.8/87.6 \\
Chicken  & 64.0/55.9 & 69.7/53.9 & 69.7/53.9 & \textbf{85.2}/67.6 & \underline{78.6}/75.9 & 78.0/- & 71.4/\textbf{79.8} & 68.6/- & 71.5/64.0 & 68.8/62.4 & 74.2/\underline{78.5} \\
Diamond  & 64.0/75.3 & 61.3/50.3 & 79.2/51.7 & 90.0/83.5 & 93.2/86.2 & 90.5/- & \underline{94.8}/92.6 & 80.1/- & \textbf{95.5}/\underline{94.2} & 71.2/76.8 & 76.3/\textbf{95.3} \\
Duck     & 55.4/71.9 & 80.5/50.3 & 80.5/56.1 & 58.4/50.3 & 67.9/63.1 & 51.7/- & 71.2/\underline{87.6} & 82.0/- & 83.1/82.2 & \textbf{94.5}/66.3 & \underline{84.8}/\textbf{90.5} \\
Fish     & 84.0/\textbf{98.8} & 67.4/51.6 & 73.4/60.2 & 91.5/82.6 & \underline{97.6}/83.6 & 88.0/- & 94.5/88.6 & 85.9/- & 86.5/93.2 & 69.5/94.2 & \textbf{100}/\underline{97.0} \\
Gemstone & 34.9/44.9 & 46.7/49.1 & 63.2/57.9 & 41.7/54.5 & 53.9/56.4 & 67.4/- & 46.8/\textbf{85.7} & \underline{69.3}/- & 56.0/45.8 & 68.8/75.0 & \textbf{72.5}/\underline{85.4} \\
Seahorse & 84.3/\textbf{96.2} & 63.8/44.5 & 71.5/50.8 & 76.2/81.7 & 84.1/82.7 & 60.4/- & 72.9/81.3 & 75.6/- & 71.6/65.9 & \textbf{92.4}/81.5 & \underline{90.9}/\underline{83.1} \\
Shell    & 39.3/72.5 & 64.7/50.2 & \textbf{81.6}/56.2 & 58.3/\underline{81.1} & 58.5/79.8 & 66.5/- & 62.3/\textbf{83.9} & 80.0/- & \underline{80.3}/77.8 & 73.3/60.2 & 69.6/63.3 \\
Starfish & 52.6/\textbf{80.0} & 68.6/45.1 & \textbf{79.4}/54.5 & 50.6/61.7 & 56.2/62.5 & 67.4/- & 66.0/64.1 & 75.8/- & \underline{76.6}/\underline{69.0} & 74.8/67.5 & 68.8/60.4 \\
Toffees  & \underline{84.5}/\underline{95.9} & 65.6/44.2 & 75.8/51.6 & 68.5/75.9 & 79.6/80.3 & 77.4/- & 84.2/89.5 & 77.1/- & 73.8/93.4 & 76.3/93.5 & \textbf{95.0}/\textbf{97.0} \\ \hline
Mean     & 62.5/75.9 & 66.6/52.0 & 75.2/56.5 & 70.4/70.5 & 75.1/73.5 & 72.5/- & 76.7/\underline{83.6} & 76.5/- & \underline{78.2}/76.8 & 75.0/76.7 & \textbf{80.9}/\textbf{84.8} \\ \bottomrule[1.5pt]

\end{tabular}}}
\end{table*}

\begin{table*}[t]
\centering
\caption{\textbf{Quantitative Results on MulSen-AD.} The results are presented in O-ROC\%/P-ROC\%. The best performance is in \textbf{bold}, and the second best is \underline{underlined}.}
\label{table:mulsen}
\fontsize{10}{14}\selectfont{
\resizebox{\linewidth}{!}{
\begin{tabular}{c|ccccc
>{\columncolor{blue!8}}c
cc|
>{\columncolor{blue!8}}c}
\toprule[1.5pt]

Method~$\rightarrow$   & PatchCore-FP & PatchCore-FP-R & PatchCore-PM & Reg3D-AD & R3D-AD & R3D-AD & IMRNet & GLFM & \textbf{\ourmethod} \\
Category~$\downarrow$  & CVPR'22 & CVPR'22 & CVPR'22 & NeurIPS'23 & ECCV'24 & +MPAS & CVPR'24 & TASE'25 & \textbf{Ours} \\ \midrule
Capsule          & 89.8/91.7 & 90.5/91.9 & 90.3/\underline{92.1} & \underline{91.2}/87.7 & 78.1/59.8 & 78.1/59.8 & 60.1/42.3 & \textbf{96.7}/\textbf{93.0} & \underline{91.2}/86.0 \\
Cotton           & 25.3/55.4 & 26.3/54.6 & 19.7/52.8 & 43.0/52.1 & 79.0/38.3 & \underline{87.5}/45.9 & 58.5/50.7 & 81.2/\underline{67.9} & \textbf{99.8}/\textbf{75.8} \\
Cube             & 72.3/57.5 & 66.8/43.7 & 72.2/41.7 & 56.9/62.6 & 73.4/60.5 & 73.4/60.5 & 43.2/56.6 & \underline{75.6}/\underline{68.0} & \textbf{90.2}/\textbf{80.5} \\
Spring pad       & 98.6/62.9 & \textbf{100}/60.1 & 96.5/62.1 & 95.1/\textbf{80.2} & 44.4/49.5 & 82.6/59.7 & 65.1/40.1 & \textbf{100}/70.5 & \textbf{100}/\underline{77.7} \\
Screw            & 97.9/57.8 & 93.1/61.0 & \textbf{99.7}/59.7 & 97.2/54.0 & 56.2/49.7 & 80.2/\underline{61.3} & 74.2/45.6 & 63.6/60.9 & \underline{99.3}/\textbf{66.3} \\
Screen           & 91.6/\textbf{60.9} & 95.0/58.7 & 89.7/53.2 & 64.1/46.6 & 91.3/56.7 & \underline{95.3}/56.7 & 37.8/35.2 & 86.6/50.8 & \textbf{99.4}/\underline{60.0} \\
Piggy            & \textbf{100}/\underline{84.8} & 99.7/62.4 & 98.2/60.3 & 86.6/63.5 & 97.2/54.8 & \textbf{100}/54.8 & 72.9/51.2 & 73.2/77.8 & 98.0/\textbf{85.8} \\
Nut              & 97.1/\underline{90.3} & \textbf{98.9}/89.6 & \textbf{98.9}/89.7 & 79.7/80.7 & 44.0/49.6 & 86.9/67.4 & 81.2/36.9 & 94.0/\textbf{95.7} & 96.6/88.7 \\
Flat pad         & \textbf{100}/70.7 & 89.3/67.8 & 94.4/63.0 & 90.8/69.2 & 87.0/60.5 & 91.3/64.5 & 71.4/54.2 & \underline{94.6}/\textbf{77.2} & \underline{94.6}/\underline{74.3} \\
Plastic cylinder & \textbf{94.1}/\textbf{83.0} & 90.8/76.6 & \underline{93.6}/\underline{76.9} & 76.5/67.0 & 56.9/57.6 & 74.5/63.9 & 62.1/41.2 & 81.5/67.8 & 81.5/71.2 \\
Zipper           & 79.7/55.2 & 81.3/54.5 & 73.9/50.2 & 47.0/53.6 & 85.7/54.1 & \underline{87.1}/54.1 & 63.0/49.6 & 81.3/\underline{57.4} & \textbf{94.8}/\textbf{61.5} \\
Button cell      & \textbf{91.5}/38.2 & 68.7/51.2 & \underline{79.7}/47.8 & 78.2/\underline{70.6} & 62.8/49.8 & 68.7/54.5 & 70.2/48.5 & 56.7/43.3 & 73.1/\textbf{75.4} \\
Toothbrush       & \textbf{90.5}/\underline{60.5} & 88.8/60.4 & \underline{89.1}/\textbf{60.6} & 81.2/47.2 & 78.0/57.0 & 81.3/58.5 & 61.5/51.9 & 84.9/57.8 & 85.9/40.9 \\
Solar panel      & 62.4/20.2 & 60.5/26.5 & 61.2/27.4 & \textbf{66.0}/60.9 & 50.2/55.1 & \underline{63.9}/\underline{61.4} & 34.4/53.3 & 40.9/57.5 & 49.3/\textbf{67.8} \\
Light            & 97.5/\textbf{70.7} & \textbf{100}/\underline{70.6} & \underline{99.2}/69.6 & 89.7/65.1 & 38.5/49.6 & 64.8/57.4 & 45.7/41.5 & 66.1/52.6 & 90.9/67.4 \\ \hline
Mean             & \underline{86.0}/64.0 & 83.3/62.0 & 84.0/60.5 & 74.9/64.1 & 68.2/53.5 & 81.0/58.7 & 60.1/46.7 & 78.5/\underline{66.5} & \textbf{89.6}/\textbf{72.0} \\ \bottomrule[1.5pt]
\end{tabular}}}
\end{table*}

\begin{figure*}[h!]
\centering\includegraphics[width=\linewidth]{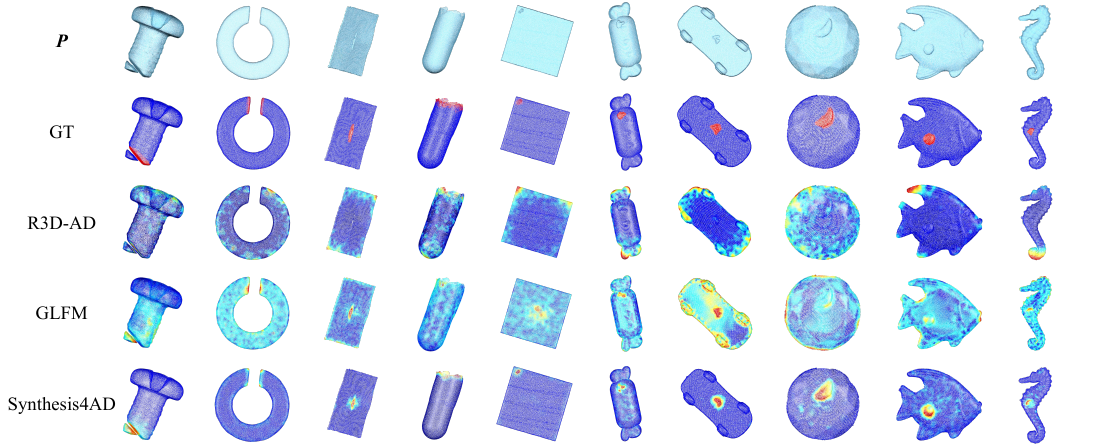}
\caption{Qualitative comparison of on representative categories. From top to bottom: input point clouds (\textbf{\textit{P}}), ground-truth masks (GT), and predicted anomaly maps produced by R3D-AD, GLFM, and our \ourmethod{}. Warmer colors indicate higher anomaly scores.}

\label{fig:Vis_results}
\end{figure*}

Table~\ref{table:real} summarizes the quantitative results on Real3D-AD in terms of O-ROC/P-ROC. Overall, \ourmethod{} achieves the best mean performance with 80.9\% O-ROC and 84.8\% P-ROC, indicating superior capability compared with competing methods. A key observation is that \ourmethod{} delivers robust detection across a wide range of categories: it reaches near-saturated accuracy on several classes with relatively clear abnormal patterns (e.g., Car: 96.3\%/95.7\%, Fish: 100\%/97.0\%, Toffees: 95.0\%/97.0\%), while still maintaining high P-ROC on categories where defects are subtle or spatially extended (e.g., Diamond: 95.3\%, Duck: 90.5\%). The effectiveness of our synthesis component is further evidenced by the comparison between R3D-AD with and without MPAS. When simply replacing the original synthesis with MPAS, the mean performance improves markedly from 66.6\%/52.0\% to 75.2\%/56.5\%, demonstrating that MPAS provides higher-fidelity and more diverse abnormal patterns that better match real defect statistics, even without altering the downstream learning paradigm. Moreover, by incorporating richer data augmentation to increase data diversity and to reduce sensitivity to absolute coordinates, \ourmethod{} learns more robust feature representations compared with GLFM, leading to a significant performance improvement.

A similar trend is observed on MulSen-AD, as shown in Table~\ref{table:mulsen}, where \ourmethod{} delivers the best mean performance of 89.6\% O-ROC and 72.0\% P-ROC, clearly exceeding suboptimal PatchCore-FP and GLFM. 
\ourmethod{} substantially boosts performance on Cotton (99.8\%/75.8\%) and Cube (90.2\%/80.5\%), while remaining competitive on categories that already exhibit high separability (e.g., Spring pad: 100\%/77.7\%, Piggy: 98.0\%/85.8\%). In addition, the ablated R3D-AD also benefits noticeably when its original synthesis is replaced by MPAS (From 68.2\%/53.5\% to 81.0\%/58.7\%), indicating that higher-fidelity and more diverse synthetic defects improve the alignment with real anomaly statistics even under the same downstream pipeline. Overall, these results suggest that the combination of realistic, geometry-consistent synthesis and mask-supervised representation learning enables \ourmethod{} to produce more reliable point-wise anomaly responses.

The qualitative results further corroborate the quantitative gains. As shown in Fig.\ref{fig:Vis_results}, R3D-AD and GLFM often produce dispersed responses and spurious activations over non-defective regions, especially on thin structures and smoothly curved surfaces, which can inflate false positives and blur defect boundaries. In contrast, the anomaly maps produced by \ourmethod{} are more compact and mask-aligned, highlighting the true defective regions with higher contrast while suppressing irrelevant background responses.

\begin{figure}[h!]
\centering
\includegraphics[width=\linewidth]{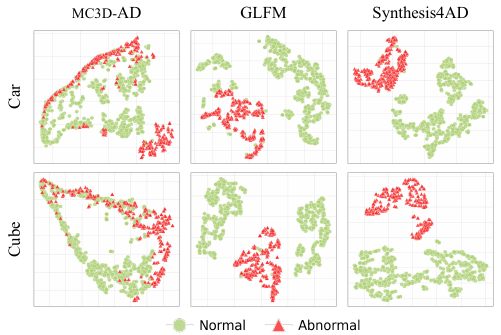} 
\caption{Feature distribution visualization of MC3D-AD, GLFM, and \ourmethod{}. All methods utilize the identical PointMAE backbone.}
\label{fig:feature_vis}
\end{figure}

Furthermore, as visualized in Fig.~\ref{fig:feature_vis}, \ourmethod{} achieves significantly clearer feature separation than MC3D-AD~\cite{MC3D} and GLFM~\cite{GLFM} under the identical PointMAE~\cite{pointmae} backbone. Unlike MC3D-AD's frozen encoder and GLFM's fine-tuning limited by simple heuristic defects, \ourmethod{} fine-tunes the network using diverse, physics-isomorphic anomalies. This strategy explicitly injects richer structural defect knowledge, yielding a highly decoupled feature space that directly drives our precise anomaly localization.

\subsection{Ablation Study}

We conduct ablation studies to quantify the impact of (i) the anomaly types used for training, (ii) the amount of synthesized anomalies, and (iii) the proposed spatial-distribution normalization (SDN) and geometry-faithful augmentations. All results are validated on Real3D-AD and MulSen-AD.

\subsubsection{Impact of anomaly types}

\begin{table}[t]
\centering
\caption{\textbf{Ablation results on the impact of different anomaly types used during training.}}
\label{table:Ablation_Types}
\fontsize{10}{12}\selectfont

\resizebox{\linewidth}{!}{

\begin{tabular}{ccc|c|c}
\toprule[1.5pt]

MPAS-1D & MPAS-2D & MPAS-3D & Real3D-AD & MulSen-AD \\
\midrule

\ding{51} & \ding{55} & \ding{55} &  75.6/79.4 & 81.3/68.3 \\
\ding{55} & \ding{51} & \ding{55} &  73.3/76.2 & 80.8/65.1 \\
\ding{55} & \ding{55} & \ding{51} &  77.6/81.8 & 84.6/69.9 \\
\ding{51} & \ding{51} & \ding{55} &  77.2/80.5 & 83.7/69.4 \\
\ding{51} & \ding{55} & \ding{51} &  78.4/83.2 & 86.2/70.8 \\
\rowcolor{blue!8} 
\ding{51} & \ding{51} & \ding{51} &  \textbf{80.9/84.8} & \textbf{89.6/72.0} \\

\bottomrule[1.5pt]
\end{tabular}
}
\vspace{-3mm}
\end{table}

Table~\ref{table:Ablation_Types} investigates how different anomaly supports affect performance, where MPAS-1D/2D/3D correspond to anomalies generated from MPAS utilizing 1D/2D/3D primitives, respectively. Only employing MPAS-1D yields limited gains, as the synthesized defects are largely localized and regular. Training with anomalies generated by MPAS-3D alone consistently outperforms other settings (MPAS-1D and MPAS-2D), indicating that volumetric supports better capture structurally coherent deformations. Moreover, combining anomaly types leads to further improvements, suggesting strong complementarity across geometric extents. In particular, enabling all three types achieves the best results (80.9\%/84.8\% on Real3D-AD and 89.6\%/72.0\% on MulSen-AD), improving over the best single-type setting by +3.3\%/+3.0\% and +5.0\%/+2.1\% in O-ROC/P-ROC, respectively. These observations validate our core motivation that higher-dimensional supports substantially expand the expressive space of plausible 3D defects and provide more informative supervision for representation learning.

\subsubsection{Impact of sample number}

\begin{figure}[h!]
\centering\includegraphics[width=\linewidth]{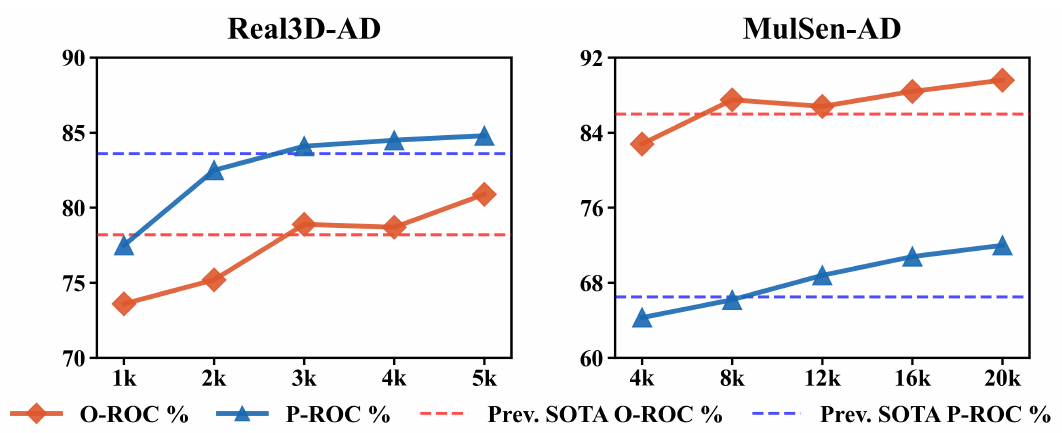}
\caption{Effect of synthetic anomaly scale on detection performance. O-ROC and P-ROC on Real3D-AD and MulSen-AD are reported as the number of synthesized anomalous training samples increases. The dashed lines indicate the previous SOTA results.}
\label{fig:ablation_scale}
\end{figure}

Further study results on the effect of the synthetic anomaly scale are reported in Fig.~\ref{fig:ablation_scale}. Overall, increasing the number of synthesized anomalies yields a clear performance improvement on both Real3D-AD and MulSen-AD, exhibiting an approximately monotonic upward trend. On Real3D-AD, O-ROC and P-ROC consistently increase as the training set grows from 1k to 5k, with the largest gain observed in the low-data regime from 1k to 2k. This suggests that even a modest increase in synthetic anomalies can substantially enhance defect diversity, quickly strengthening the supervision signal and encouraging the model to learn local geometric features more strongly correlated with anomalous patterns. On MulSen-AD, performance similarly improves as the anomaly count increases from 4k to 20k. While slight fluctuations appear at intermediate budgets—likely due to stochastic variations in the synthesized defect distribution—the overall trajectory remains upward and steadily exceeds the previous SOTA baselines (dashed lines). Collectively, these results corroborate our key claim that large-scale, high-fidelity anomaly synthesis offers scalable and effective supervision, improving both detection accuracy and generalization.

\subsubsection{Impact of SDN and Data Augmentation}

Table~\ref{table:ablation_aug} reports the ablation results of Spatial-Distribution Normalization (SDN) and three augmentations (random rotations, noise perturbations, and point dropout) on Real3D-AD and MulSen-AD. 
Compared with the baseline without SDN and augmentations, introducing SDN alone yields a substantial improvement on both benchmarks, boosting Real3D-AD from 74.7\%/77.5\% to 78.3\%/82.0\% and MulSen-AD from 82.4\%/65.2\% to 86.5\%/68.8\%. This confirms that scale-aligned normalization effectively regularizes point cloud statistics and stabilizes training across category-dependent scales and densities. 
Building upon SDN, adding random rotations consistently improves performance, with 79.0\%/83.1\% on Real3D-AD and 87.4\%/70.7\% on MulSen-AD, indicating enhanced robustness to pose variations and reduced sensitivity to absolute coordinates. 
Further incorporating noise perturbations brings additional gains, suggesting that modeling acquisition noise helps the encoder learn more transferable local geometric information. 
Point dropout also contributes complementary benefits by simulating partial observations, which is especially relevant for real scanning artifacts.
Combining SDN with all three augmentations, \ourmethod{} achieves the best performance, reaching 80.9\%/84.8\% on Real3D-AD and 89.6\%/72.0\% on MulSen-AD, demonstrating that SDN and the proposed augmentations are mutually reinforcing for improving both effectiveness and generalization.

\begin{table}[t]
\centering
\caption{\textbf{Ablation results on the impact of SDN and different data augmentation.}}
\label{table:ablation_aug}
\fontsize{10}{12}\selectfont

\resizebox{\linewidth}{!}{

\begin{tabular}{cccc|c|c}
\toprule[1.5pt]

SDN & Rotations & Perturbations & Dropout & Real3D-AD & MulSen-AD \\
\midrule

\ding{55} & \ding{55} & \ding{55} & \ding{55} & 74.7/77.5 & 82.4/65.2 \\
\ding{51} & \ding{55} & \ding{55} & \ding{55} & 78.3/82.0 & 86.5/68.8 \\
\ding{51} & \ding{51} & \ding{55} & \ding{55} & 79.0/83.1 & 87.4/70.7 \\
\ding{51} & \ding{51} & \ding{51} & \ding{55} & 80.1/84.2 & 88.9/71.4 \\
\ding{51} & \ding{51} & \ding{55} & \ding{51} & 79.6/83.3 & 87.7/71.2 \\
\rowcolor{blue!8} 
\ding{51} &  \ding{51} &  \ding{51} & \ding{51} & \textbf{80.9/84.8} & \textbf{89.6/72.0} \\

\bottomrule[1.5pt]
\end{tabular}
}
\vspace{-3mm}
\end{table}

\begin{figure}[h!]
\centering
\includegraphics[width=\linewidth]{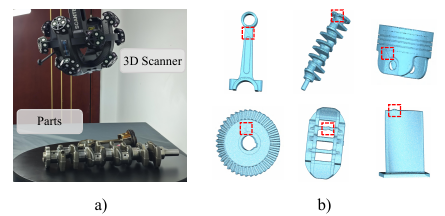}
\caption{Real-world industrial data collection and representative defect examples. (a) The 3D scanning setup is utilized to capture point clouds of industrial parts. (b) Abnormal samples from six object categories; red boxes highlight defective position.}

\label{fig:realdata}
\end{figure}

\begin{table}[t]
\centering
\caption{\textbf{Quantitative Results on Actual Industry Parts Dataset.} The results are presented in O-ROC\%/P-ROC\%. The best performance is in \textbf{bold}.}
\label{table:real_collected}
\fontsize{10}{14}\selectfont{

\resizebox{\columnwidth}{!}{
\begin{tabular}{l|ccc
>{\columncolor{blue!8}}c} 
\toprule[1.5pt]

Method~$\rightarrow$   & R3D-AD & MC3D-AD & GLFM & \textbf{\ourmethod} \\
Category~$\downarrow$  & ECCV'24 & IJCAI'25 & TASE'25 & \textbf{Ours} \\ \midrule

Bevel Gear     & 63.7/42.5 & 80.7/62.1 & 83.7/64.1 & \textbf{100}/\textbf{72.1} \\
Brake Caliper  & 64.0/47.7 & 83.0/65.6 & 83.0/66.6 & \textbf{93.7}/\textbf{70.1} \\
Connecting Rod & 56.8/54.2 & 80.0/63.7 & 68.0/66.3 & \textbf{97.3}/\textbf{72.9} \\
Crankshaft      & 44.3/53.1 & 91.3/69.5 & 81.3/65.8 & \textbf{87.7}/\textbf{69.1} \\
Piston          & 65.3/60.0 & 79.7/61.2 & 82.3/66.7 & \textbf{96.7}/\textbf{75.9} \\
Turbine Blade  & 71.0/51.6 & 73.3/60.6 & 98.7/65.2 & \textbf{100}/\textbf{82.9} \\ \hline

Mean            & 60.9/51.5 & 81.3/63.8 & 82.8/65.8 & \textbf{95.9}/\textbf{73.8} \\ \bottomrule[1.5pt]

\end{tabular}}}
\end{table}

\begin{figure}[h!]
\centering
\includegraphics[width=\linewidth]{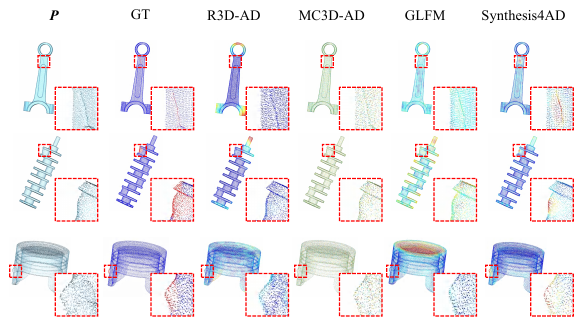}
\caption{\textbf{Visualization of prediction results in the actual industry parts dataset using the proposed method and other methods.} The first column is the original point clouds, while the second column is the ground truth. Subsequent columns depict various methods.}
\label{fig:realvis}
\end{figure}

\subsection{Evaluation in Practical Industrial Parts}

To evaluate the practicality of our system in real inspection scenarios, we build a real-world dataset by scanning physical industrial parts with a 3D scanner, as illustrated in Fig.~\ref{fig:realdata} (a). The dataset covers six representative object categories---Bevel Gear, Brake Caliper, Connecting Rod, Crankshaft, Piston, and Turbine Blade---with typical manufacturing defects highlighted in Fig.~\ref{fig:realdata} (b). Following the standard unsupervised protocol, the training split contains only normal samples, consisting of 20 normal point clouds per category. The test split includes both normal and defective samples, with 10 normal and 30 abnormal point clouds, respectively. 

Table~\ref{table:real_collected} reports the quantitative results on the collected dataset.  Overall, \ourmethod{} achieves a clear margin over all competing methods, reaching 95.9\%/73.8\% on average. In particular, compared with the previous SOTA GLFM, our method \ourmethod{} improves the mean performance by +13.1\% O-ROC and +8.0\% P-ROC, indicating substantially stronger capability in both object-wise and point-wise detection under real scanning conditions.  Notably, \ourmethod{} attains near-saturated object-wise detection on several categories, such as Bevel Gear and Turbine Blade with 100\% O-ROC, while consistently maintaining higher P-ROC across all categories. These gains suggest that training with large-scale, high-fidelity synthetic anomalies leads to more discriminative features.

Qualitative comparisons are further provided in Fig.~\ref{fig:realvis}, where columns show the input point clouds (P), ground-truth masks (GT), and predicted anomaly maps from different methods. These baseline approaches tend to produce either diffuse responses or spurious activations on normal regions, resulting in false positives and fragmented localization. In contrast, \ourmethod{} yields more concentrated and geometry-aligned anomaly responses that better coincide with the annotated defect regions, while suppressing background noise on normal surfaces. This qualitative evidence is consistent with the quantitative improvements in P-ROC, demonstrating that our model not only detects defective objects reliably but also localizes subtle defects more precisely in real industrial scans.

\section{Conclusion}\label{sec:conclusion}

In this work, we address the data scarcity and long-tail challenge in industrial 3D anomaly detection by constructing a unified system that couples high-fidelity anomaly synthesis with downstream detector learning and deployment. We first introduce Multi-dimensional Primitive-guided Anomaly Synthesis (MPAS), a high-fidelity 3D anomaly synthesis framework that generates diverse and geometrically realistic defects together with accurate point-wise anomaly masks. Built upon MPAS, we further develop 3D-DefectStudio, which provides both an interactive interface and a programmable Python API, making controllable and large-scale synthetic data generation practical for industrial use.
On this basis, we propose \textbf{Synthesis4AD}, an end-to-end workflow that integrates knowledge-driven anomaly generation, supervised detector training, and prototype-based online inference within a single framework. Specifically, Synthesis4AD leverages a multimodal large language model (MLLM) to parse product design information, expert priors, and multimodal specifications into executable synthesis instructions, which drive 3D-DefectStudio to generate anomalies that are better aligned with defects encountered in manufacturing and real usage. The resulting synthetic anomaly--mask pairs are then used to train the downstream detector, where spatial-distribution normalization and training-time data augmentations improve feature robustness and cross-category generalization. During deployment, the trained encoder is further combined with prototype-based feature matching to produce both object-level anomaly scores and point-wise localization results.
Extensive experiments on public benchmarks, comprehensive ablation studies, and real-world inspections on industrial parts consistently demonstrate the effectiveness, robustness, and generalization ability of the proposed system. The results show that Synthesis4AD provides a scalable and practical route toward reliable 3D anomaly detection, achieving state-of-the-art performance in both object-level detection and fine-grained point-wise localization.

A current limitation of Synthesis4AD is that its workflow remains open-loop: the quality of synthesized anomalies is determined before detector training and is not further refined based on downstream performance. In future work, we plan to close this loop by feeding quantitative and qualitative feedback from training and inference back into the MLLM-driven synthesis module, enabling adaptive refinement of the generated anomalies. Such a closed-loop optimization process may further improve the quality of synthetic supervision, strengthen learned representations, and ultimately yield more robust deployment performance.

\ifCLASSOPTIONcaptionsoff
  \newpage
\fi



%

{\small
\bibliographystyle{unsrt}

\bibliography{ref}
}
%

\end{document}